\documentclass[runningheads]{llncs}
\usepackage{subfig}
\usepackage{graphicx}
\usepackage{amsmath}

\usepackage{booktabs}
\usepackage[normalem]{ulem}
\useunder{\uline}{\ul}{}
\usepackage{multirow}

\usepackage{tcolorbox}

\usepackage{orcidlink}

\begin{document}
\title{Beyond Self-Consistency: Ensemble Reasoning Boosts Consistency and Accuracy of LLMs in Cancer Staging\thanks{This manuscript has been accepted to the 22nd International Conference on Artificial Intelligence in Medicine (AIME'24).}}
\titlerunning{Ensemble Reasoning in Cancer Staging}
%
\author{Chia-Hsuan Chang\inst{1}\orcidlink{0000-0001-9116-8244} \and Mary M. Lucas\inst{1}\orcidlink{0000-0002-0413-7499} \and
Yeawon Lee\inst{1}\orcidlink{0009-0009-4209-2672} \and
Christopher C. Yang\inst{1}\orcidlink{0000-0001-5463-6926} \and
Grace Lu-Yao\inst{2}\orcidlink{0000-0002-2925-7737}}
\authorrunning{Chang et al.}

\institute{College of Computing \& Informatics, Drexel University, Philadelphia PA, USA \\
\email{\{cc3859,mml367,yl3427,chris.yang\}@drexel.edu} \and
Department of Medical Oncology, Sidney Kimmel Cancer Center, \\ Thomas Jefferson University, Philadelphia, PA, USA \\
\email{Grace.LuYao@jefferson.edu}}

\maketitle              
\begin{abstract}
Advances in large language models (LLMs) have encouraged their adoption in the healthcare domain where vital clinical information is often contained in unstructured notes. Cancer staging status is available in clinical reports, but it requires natural language processing to extract the status from the unstructured text. With the advance in clinical-oriented LLMs, it is promising to extract such status without extensive efforts in training the algorithms. Prompting approaches of the pre-trained LLMs that elicit a model's reasoning process, such as chain-of-thought, may help to improve the trustworthiness of the generated responses. Using self-consistency further improves model performance, but often results in inconsistent generations across the multiple reasoning paths. In this study, we propose an ensemble reasoning approach with the aim of improving the consistency of the model generations. Using an open access clinical large language model to determine the pathologic cancer stage from real-world pathology reports, we show that the ensemble reasoning approach is able to improve both the consistency and performance of the LLM in determining cancer stage, thereby demonstrating the potential to use these models in clinical or other domains where reliability and trustworthiness are critical.

\end{abstract}

\section{Introduction}

The standardized TNM cancer staging system is an essential part of cancer diagnosis and management, classifying the extent of cancer based on tumor size (T), lymph node involvement (N), and metastasis (M). Pathology reports detail this information based on tissue sample analyses to determine the pathologic TNM (pTNM) stage, but their free-text format complicates large-scale, rapid data extraction. Natural Language Processing (NLP) technologies, especially generative Large Language Models (LLMs), show promise in automating cancer staging from these reports. These models, capable of zero-shot (ZS) and few-shot (FS) learning, circumvent the need for extensive and costly human-annotated training data, which has been a limiting factor in the application of these technologies to real-world clinical settings.

To effectively induce and maximize the emergent learning capabilities, LLMs can benefit from various sophisticated prompt engineering techniques. For example, the Chain-of-Though approach~(CoT)~\cite{wei_chain--thought_2023,kojima_large_2023} has proven effective by eliciting the reasoning process behind the model’s predictions. This approach can enhance the interpretability of these predictions and allow for a degree of traceability in the model’s reasoning path and the source of its responses. However, LLMs' inherent stochastic generation process can lead to varying reasoning paths and inconsistent responses, a critical issue in clinical scenarios where accountability is crucial~\cite{sivarajkumar_empirical_2023}.

To mitigate these challenges, we develop an Ensemble-Reasoning (EnsReas) approach, building upon the concept of self-consistency (SC) ~\cite{wangSelfconsistencyImprovesChain2023}. EnsReas enhances LLM analysis of pathology reports by first leveraging self-consistency as an intermediate step to generate multiple reasoning and prediction responses. These reasonings are then grouped by predicted outcomes, forming a revised prompt that guides the LLM to re-assess inconsistent initial answers, ensuring a more robust and reliable assessment. Our results demonstrate that EnsReas performs better than the baseline methods (ZS and ZS-CoT) regarding predictive performance. Moreover, it outperforms ZS-CoT-SC in predictive performance and consistency, suggesting that LLMs can enhance their decision-making process using the EnsReas approach, leading to more consistent and reliable responses for complex tasks like determining cancer staging from pathology reports. This advancement underscores LLMs' potential in interpreting and utilizing clinical data for effective cancer treatment planning.

\section{Related Works}

The use of NLP has become increasingly popular for extracting valuable information from pathology reports. Researchers have proposed various approaches to classify pathologic stages, including traditional machine learning models~\cite{odishoNaturalLanguageProcessing2020} such as logistic regression, AdaBoost, and random forest, as well as specific neural network architectures. Hierarchical networks that learn representation from words to sentences and reports have been proposed in previous studies~\cite{gaoHierarchicalAttentionNetworks2018,gaoClassifyingCancerPathology2019}, while others~\cite{wuStructuredInformationExtraction2020} have experimented with the attention-based graph convolution network. Additionally, fine-tuning clinical-specific models, such as Clinical-BigBird, has been shown to be effective for TNM classification~\cite{kefeli_generalizable_2023}. 
Our previous work~\cite{changClassifyingCancerStage2024} demonstrated that open-source clinical LLMs can reach similar and even superior predictive performance compared with the supervised model on TNM tasks. However, this previous work only evaluated the existing prompting strategies such as zero-shot, few-shot and zero-shot chain-of-thought. 

In order to maximize the potential of LLMs, various prompting approaches have been proposed. One stable method is the Chain-of-Thought (CoT) approach, introduced by Wei et al.~\cite{wei_chain--thought_2023}, which mimics the human thought process in problem-solving. In this method, the model is provided with a few examples that includes a series of reasoning steps leading to the final answer. To address the challenge of creating specific demonstrations for each task, Kojima et al.~\cite{kojima_large_2023} proposed the zero-shot CoT (ZS-CoT) approach. This method involves appending a simple prompt that induces a reasoning process, such as `Let's think step by step', without requiring specific examples.

Although effective, the CoT method can be further refined considering the inherent characteristics of LLMs, which include elements of randomness. 
A single sampling can result in flawed reasoning and potentially inaccurate answer. Additionally, even subtle modifications to the prompts can result in significant variations in model's predictions, thus increasing the variability of the output~\cite{sivarajkumar_empirical_2023}. To address this challenge, Wang et al.~\cite{wangSelfconsistencyImprovesChain2023} proposed the Self-Consistency approach, which involves sampling a diverse set of reasoning paths and aggregating the final answer by marginalizing out these paths. This helps in finding the most consistent answer among multiple reasoning paths. Building on the concept of CoT, Karan et al.~\cite{singhal_towards_2023}, integrate CoT and Self-Consistency with an ensemble refinement (ER) to suggest improvement for medical question answering task. While ER improves response quality the authors acknowledge the increased resource costs associated with repeated samplings and recommend applying these methods selectively to certain types of questions.

In this work, we propose the ensemble reasoning (EnsReas) approach to perform cancer staging task in pathologic reports. Compared to ER, EnsReas has two contributions: (1) EnsReas relies on ZS-CoT, thus eliminating the need for manual-compiled CoT that is required for ER, and (2) unlike ER which repeats sampling process on all predictions, EnsReas first separates consistent and inconsistent predictions and only re-assesses the inconsistent predictions, making it a more cost-effective approach.  

\section{Materials and Methods}

\subsection{Data}
We use a real-world corpus of breast cancer pathology reports from the Cancer Genomic Atlas (TCGA) project. The raw reports, in PDF format, are available from the National Cancer Institute (NCI) Genomic Data Commmons (GDC) portal. We utilize a preprocessed subset of reports curated and described in \cite{kefeliTCGAReportsMachinereadablePathology2024} and made available for download\footnote{https://github.com/tatonetti-lab/tcga-path-reports}. We focus our experiments on predicting pathologic T and N stage for breast cancer because it is one of the top diagnosed cancers in the United States\footnote{https://seer.cancer.gov/statfacts/html/common.html} and has good representation of reports and ground truths in the dataset. Because some of the pathology reports do not report all pTNM stages within the same report, we follow \cite{kefeli_generalizable_2023} and treat the T stage and N stage prediction as two different tasks.

\begin{table}[]
\centering
\caption{Distribution of class for T and N category}
\label{tab:distribution-of-class}
\resizebox{0.55\textwidth}{!}{%
\begin{tabular}{@{}llllll@{}}
\toprule
\multirow{2}{*}{N Ctategory} & N0 & N1 & N2 & N3 & Total \\
 & 316 & 300 & 110 & 74 & 800 \\
\multirow{2}{*}{T Category} & T1 & T2 & T3 & T4 & Total \\
 & 589 & 273 & 131 & 38 & 1031 \\ \bottomrule
\end{tabular}%
}
\end{table}

\subsection{Language Model}

We utilize Med42-70B\footnote{https://huggingface.co/m42-health/med42-70b}, an open-access clinical LLM, installed on a local server. Med42-70B is derived from Llama2-70B and instruction-tuned on a dataset of medical knowledge, with reported superior performance in the zero shot setting compared to GPT-3.5.  When compared with other open access clinical generative LLMs, Med42-70B outperforms ClinicalCamel-70B.

\subsection{Baselines}

We first implement three prompting strategies, zero shot (ZS), zero shot chain-of-thought (ZS-CoT)~\cite{kojima_large_2023}, and ZS-CoT with self-consistency (ZS-CoT-SC)~\cite{wangSelfconsistencyImprovesChain2023}, to obtain baseline performance for our task. For ZS prompting we provide the LLM with the report and instruction to return the predicted stage~$p$ for the pathologic T stage or the pathologic N stage. In ZS-CoT, we provide the report and instruct the LLM to first ``think step by step'' to retrieve the generated reasoning~$c$, and treat $c$ as the context for the LLM to predict the stage~$p$. We adopt greedy decoding for ZS and ZS-CoT to get the most likely stage prediction $p$ for each report, and we measure the performance of ZS and ZS-CoT based on their generated $p$ for each report. As for ZS-CoT-SC, for each report we adopt temperature sampling~\footnote{We set temperature as 0.7 and top p as 0.95 in our study.} on ZS-CoT to obtain 10 responses from the LLM, denoted as $(C,P)$, where $C$ is a list of 10 generated reasonings and $P$ is a list of 10 generated predicted stages. The majority vote (the most frequent answer) from $P$ is considered to be the final prediction for a report, and is used to measure performance.

\subsection{Ensemble-Reasoning (EnsReas)} 

EnsReas requires the outputs of ZS-CoT-SC for each report $r$. Therefore, each report $r \in R$ has a list of 10 reasonings $C$ and a list of 10 predicted answers $P$. By analyzing the $P$ for each report, all reports can be automatically separated into reports with consistent predictions~$R^{con}$ and reports with inconsistent predictions~$R^{inc}$. The $R^{con}$ are reports that have only one unique predicted stage in their $P$, and the $R^{inc}$ are a subset of reports by filtering out $R^{con}$ from all reports. Because the predictions for reports in $R^{con}$ are deterministic, the EnsReas keeps using the same predictions as ZS-CoT-SC. 

For the reports in $R^{inc}$, we design a prompt for EnsReas to simulate a panel discussion, triggering the LLM to resolve the inconsistent reasonings and predictions of a given report. Specifically, for each report with various answers, we aggregate the reasonings by each answer, yielding a set of grouped reasonings $g_p$:
    
\begin{equation} \label{eq. ensembled_reasonings}g_p = \{c  |  c \text{ is reasoning for answer } p ,c \in C, p \in P\}, \end{equation}

\noindent where $g_p$ can be considered as a set of reasonings (opinions) from experts who choose $p$ as the answer. Take a report with four different predicted stages (i.e., T1, T2, T3, and T4) as example, the following prompt demonstrates how we integrate $g_p$ in defining the prompt:

\begin{tcolorbox}[]

prompt = \verb|"""|

Report: \{report\}

\vspace{\baselineskip}
Panel Responses: 

T1: \{$g_{T1}$: reasonings that support T1 as the answer\}

T2: \{$g_{T2}$: reasonings that support T2 as the answer\}

T3: \{$g_{T3}$: reasonings that support T3 as the answer\}

T4: \{$g_{T4}$: reasonings that support T4 as the answer\}

\vspace{\baselineskip}
You are provided with the pathology report and the chosen answers from the panel of experts with the corresponding reasonings. 
The reasonings provided by the experts are aggregated by chosen answer.

\vspace{\baselineskip}
Please review each report. Analyze the reasonings provided by the panel for the chosen answers. 
Keep in mind that the majority vote may not be the correct one, therefore you should review report carefully in addition to considering the panel reasonings.

\vspace{\baselineskip}
The correct answer is 

\verb|"""|
\end{tcolorbox}

With this prompt template, we instruct the LLM to review every report in $R^{inc}$ and its grouped reasonings. To fairly compare with ZS-CoT-SC, EnsReas also adopts temperature sampling to have a set of 10 updated predictions for a report, denoted $P^{update}$. The most frequent answer in $P^{update}$ will be used for performance evaluation. Moreover, to understand how EnsReas improves those inconsistent predictions made by ZS-CoT-SC, we will compare every report's $P$ and $P^{update}$ generated by ZS-CoT-SC and EnsReas, respectively.
    
\subsection{Evaluation metrics} To measure the predictive performance of each prompting strategy, we report macro precision, macro recall, and macro F1 score, where macro average is taken for considering the performances of all possible stages $K$ in each category. 

\begin{align*}
\text{Macro Precision} &= \frac{1}{K} \sum_{k=1}^{K} \text{Precision}_k \\
\text{Macro Recall} &= \frac{1}{K} \sum_{k=1}^{K} \text{Recall}_k \\
\text{Macro F1-score} &= \frac{2 \cdot \text{Macro Precision} \cdot \text{Macro Recall}} {\text{Macro Precision} + \text{Macro Recall}}
\end{align*}

We use entropy to measure the consistency of predictions generated by ZS-CoT-SC and EnsReas. Specifically, given a report~$r$, we have 10 predictions~($P$) generated by ZS-CoT-SC and 10 predictions~($P^{updated}$) generated by EnsReas. We then determine the consistency by averaging the entropy of all reports: 
\begin{equation}
    \frac{\sum_{r \in R} \text{entropy}(p^r_1, ..., p^r_{10})}{R},
\end{equation}

\noindent where $p^r_i \in P, P^{update}$ and $i \in [1,10]$. A higher average entropy suggests a method has higher inconsistency in the predictions.

\section{Result and Discussion}

Table~\ref{tab:predictive-reports} presents the predictive performance of our proposed EnsReas and the two baselines. We observe that ZS has the worst performance among all strategies. By asking the language model to perform reasoning before making the prediction, ZS-CoT has a significantly increased performance in T category. While ZS-CoT has higher precision in N category, its recall drops, leading to comparable F1-score with ZS. When comparing between ZS-CoT and ZS-CoT-SC, ZS-CoT-SC only has slightly higher and comparable macro f1-score with ZS-CoT in T category and N category, respectively. The inconsistency of predictions generated by ZS-CoT-SC (see Table \ref{tab:consistency-evaluation}) may provide a reason why it cannot reach better predictive performance. Moreover, EnsReas approach performs the best in both T and N categories. These results suggest that the large language model (i.e., Med42-70B in this study) is capable of refining its generations for delivering more accurate answers in the cancer staging task. We further note that EnsReas achieves improved performances without external knowledge or human intervention.

\begin{table}[]
\centering
\caption{Predictive performance of each prompting strategy. All precision, recall, f1-score are macro average across different classes.}
\label{tab:predictive-reports}
\resizebox{\textwidth}{!}{%
\begin{tabular}{@{}lllllllll@{}}
\toprule
 & \multicolumn{4}{l}{T Category} & \multicolumn{4}{l}{N Category} \\
 & precision & recall & f1-score & support & precision & recall & f1-score & support \\ \midrule
ZS & 0.725 & 0.725 & 0.688 & 1031 & 0.848 & {\ul \textbf{0.810}} & 0.823 & 800 \\
ZS-CoT & 0.855 & 0.740 & 0.790 & 1031 & 0.873 & 0.788 & 0.825 & 800 \\
ZS-CoT-SC & {\ul \textbf{0.865}} & 0.738 & 0.793 & 1031 & 0.868 & 0.795 & 0.825 & 800 \\
EnsReas & 0.860 & {\ul \textbf{0.755}} & {\ul \textbf{0.800}} & 1031 & {\ul \textbf{0.875}} & 0.808 & {\ul \textbf{0.838}} & 800 \\ \bottomrule
\end{tabular}%
}
\end{table}

Table~\ref{tab:consistency-evaluation} reports the consistency performances on ZS-CoT-SC and EnsReas. Since EnsReas depends on ZS-CoT-SC to generate a set of reasonings as input, we consider the predictive outcomes of ZS-CoT-SC as the reference to demonstrate the changes of consistency achieved by EnsReas. We find that EnsReas generates more consistent predictions, supported by its significant lower average entropy value.

\begin{table}[]
\centering
\caption{Consistency comparison between ZS-CoT and EnsReas. The paired samples t-test is employed and *** indicates the significant difference. }
\label{tab:consistency-evaluation}
\resizebox{0.75\textwidth}{!}{%
\begin{tabular}{@{}lllll@{}}
\toprule
 &  & N & Mean & Std. Deviation \\ \midrule
\multirow{2}{*}{T Category} & ZS-CoT-SC & 1031 & 0.162 & 0.296 \\
 & EnsReas & 1031 & {\ul \textbf{0.036}}$^{***}$ & 0.147 \\
\multirow{2}{*}{N Category} & ZS-CoT-SC & 800 & 0.093 & 0.211 \\
 & EnsReas & 800 & {\ul \textbf{0.023}}$^{***}$ & 0.110 \\ \bottomrule
\end{tabular}%
}
\end{table}

\begin{figure}%
    \centering
    \subfloat[\centering T Category]{{\includegraphics[scale=0.45]{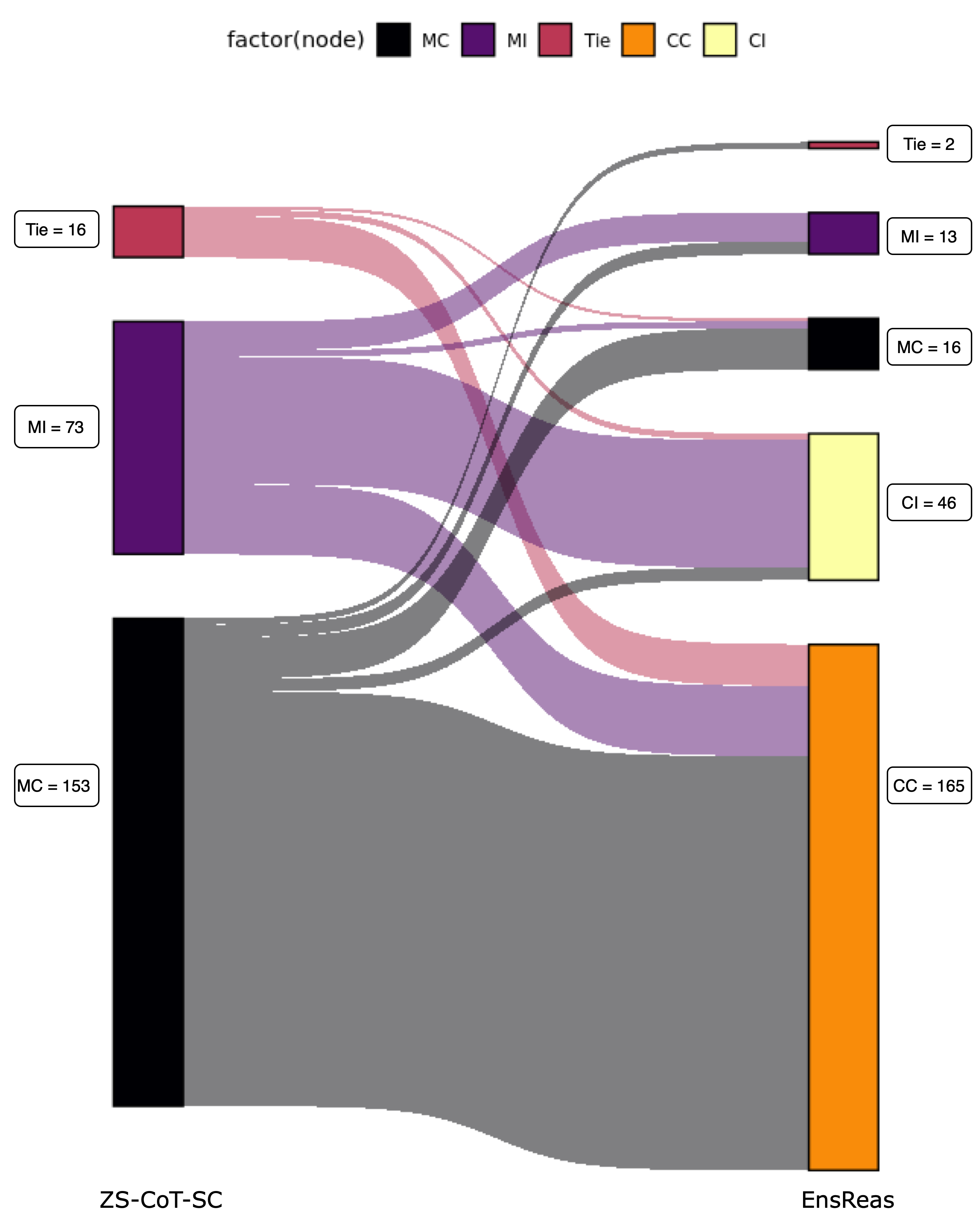} }}%
    \quad
    \subfloat[\centering N Category]
    {{\includegraphics[scale=0.45]{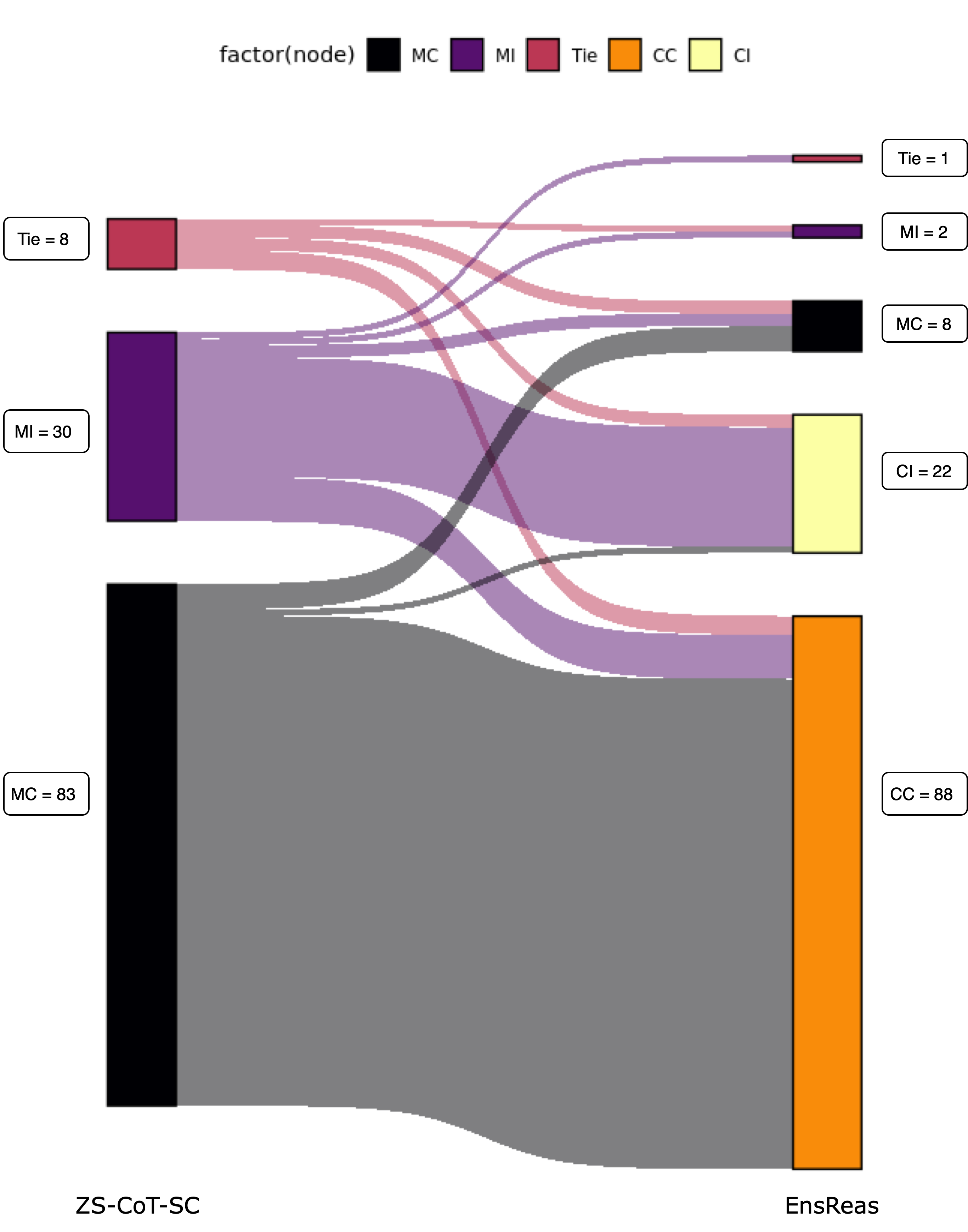} }}%
    \caption{Analysis of reasoning type between ZS-CoT-ZS and EnsReas.}%
    \label{fig:type-sequence}%
\end{figure}

To further illustrate the prediction consistency, we analyze those reports with inconsistent predictions made by ZS-CoT-SC ($P$) and compare the predictions made by EnsReas ($P^{update}$). Based on the predictions, we categorize each report into the following five different types:

\begin{enumerate}
    \item Completely correct (CC): all generated predictions for a report are correct (i.e., match the ground truth).
    \item Majority correct (MC): the majority of predictions for a report are correct with a few incorrect predictions.
    \item Tie: the number of correct predictions for a report is tied with those of an incorrect prediction.
    \item Majority incorrect (MI): the majority of predictions for a report are incorrect but there is at least one correct prediction.
    \item Completely incorrect (CI): none of the predictions for a report is correct.
\end{enumerate}

Fig.~\ref{fig:type-sequence} illustrates how EnsReas improves both predictive performance and consistency by resolving the inconsistent predictions made by ZS-CoT-SC. In the figure, we discover three patterns: (1) MC $\rightarrow$ CC: most reports with correct answer as majority becomes reports with all correct answers. (2) Tie $\rightarrow$ CC: most reports with equal number of predictions on different answers change to be CC, meaning they convert from having an undetermined answer to having a correct answer. (3) MI $\rightarrow$ CC: a noticeable portion of MI becomes CC, which shows the design of EnsReas has a chance to correct the incorrect predictions. All three patterns show how EnsReas results in more deterministic and consistent predictions. Both (2) and (3) support why EnsReas outperforms ZS-CoT-SC.

When using LLM for clinical tasks, the random inference process can be problematic~\cite{sivarajkumar_empirical_2023}. This randomness can lead to errors and inconsistencies in the output, resulting in less accurate results and generations that cannot be relied on by clinicians and other healthcare providers. Our experiments have shown that EnsReas helps to reduce the impact of randomness on the accuracy of outputs (see Table \ref{tab:predictive-reports}) and improves the consistency (Table \ref{tab:consistency-evaluation}) of LLM in cancer staging classification. This could potentially reduce clinician workload without an increased risk of misdiagnosis and inappropriate treatment.

\section{Conclusion \& Future Work}

In this work, we proposed and investigated the use of EnsReas to improve the consistency and performance of LLMs applied to a clinical task: determining cancer stage from pathology reports. Our experimental results indicate that EnsReas not only generates more accurate predictions for cancer staging but also reduces inconsistencies in LLM outputs, addressing a significant concern regarding the reliability of LLM-based predictions in clinical settings. These findings suggest that EnsReas can enhance the trustworthiness of LLMs in clinical decision-making. By mitigating randomness and improving both consistency and accuracy, EnsReas paves the way for responsible integration of LLMs into clinical workflows, potentially reducing clinician workload without introducing risks of misdiagnosis or inappropriate treatment. Future research is needed to explore the application of EnsReas to a broader range of clinical tasks, perform qualitative analysis on EnsReas's generated reasonings, and investigate mechanisms to incorporate clinician's feedback into EnsReas. Additionally, investigating methods for quantifying the confidence associated with LLM outputs could provide a crucial indicator of clinical utility.

\subsubsection{Acknowledgement.}  This work was supported in part by the National Science Foundation under the Grants IIS-1741306 and IIS-2235548, and by the Department of Defense under the Grant DoD W91XWH-05-1-023.  This material is based upon work supported by (while serving at) the National Science Foundation.  Any opinions, findings, and conclusions or recommendations expressed in this material are those of the author(s) and do not necessarily reflect the views of the National Science Foundation.

%
%
%

\bibliographystyle{splncs04}
\bibliography{main}

\end{document}